\documentclass{article}

\usepackage{times}
\usepackage{graphicx} 
\usepackage{subfigure}

\usepackage{natbib}

\usepackage{algorithm}
\usepackage{algorithmic}
\usepackage[algo2e,lined,boxed,vlined]{algorithm2e}

\usepackage{amsfonts}
\usepackage{amssymb}
\usepackage{booktabs}

\usepackage{hyperref}
\usepackage{xurl}

\usepackage[accepted]{icml2016}

\usepackage{comment}
\usepackage{amsthm}
\usepackage{mathtools}
\usepackage{tabularx}
\usepackage{multirow}

\def\b{\ensuremath\boldsymbol}

\usepackage{lastpage}
\usepackage{fancyhdr}
\usepackage{url}
\icmltitlerunning{Wittgenstein's Family Resemblance Clustering Algorithm}

\begin{document}


\twocolumn[
\icmltitle{Wittgenstein's Family Resemblance Clustering Algorithm}


\icmlauthor{Golbahar Amanpour, Benyamin Ghojogh}{}
\icmladdress{Waterloo, Ontario, Canada
\\
\\The authors contributed equally to this work.
\\Correspondence goes to {\sc bghojogh@uwaterloo.ca}
\\GitHub codes: \url{https://github.com/bghojogh/WFR_Clustering}
}


\vskip 0.3in
]

\begin{abstract}
This paper, introducing a novel method in philomatics, draws on Wittgenstein's concept of family resemblance from analytic philosophy to develop a clustering algorithm for machine learning. According to Wittgenstein's Philosophical Investigations (1953), family resemblance holds that members of a concept or category are connected by overlapping similarities rather than a single defining property. Consequently, a family of entities forms a chain of items sharing overlapping traits. This philosophical idea naturally lends itself to a graph-based approach in machine learning. Accordingly, we propose the Wittgenstein's Family Resemblance (WFR) clustering algorithm and its kernel variant, kernel WFR. This algorithm computes resemblance scores between neighboring data instances, and after thresholding these scores, a resemblance graph is constructed. The connected components of this graph define the resulting clusters. Simulations on benchmark datasets demonstrate that WFR is an effective nonlinear clustering algorithm that does not require prior knowledge of the number of clusters or assumptions about their shapes.
\end{abstract}


\section{Introduction}\label{section_introduction}

This paper introduces a novel approach in philomatics\footnote{Philomatics is a combination of philosophy and mathematics (or machine learning which is a field of mathematics). For more information, refer to \cite{ghojogh2023philomatics}.} \cite{ghojogh2023philomatics} that draws inspiration from Wittgenstein's concept of \textit{family resemblance} in analytic philosophy to develop a clustering algorithm for machine learning. 
In 1953, in his book \textit{Philosophical Investigations}, Ludwig Wittgenstein proposed that members of a concept or category are related through overlapping similarities rather than by a single defining property \cite{wittgenstein1953philosophical}. According to this view, a family of entities forms a network or chain of interrelated items, where each member resembles some, but not necessarily all, other members. This insight has been widely applied in analytic philosophy to explain the structure of concepts in language, ethics, art, and science, where traditional essentialist definitions fail.

The notion of family resemblance naturally suggests a graph-based perspective for clustering in machine learning. In a dataset, individual data points can be viewed as “members” of a conceptual family, and their pairwise resemblances (or similarities) correspond to the overlapping traits emphasized by Wittgenstein. By representing these resemblances as the edges in a graph, clusters can emerge as connected groups of points, analogous to overlapping networks of resemblance in philosophical concepts.

Motivated by this analogy, we propose the Wittgenstein's Family Resemblance (WFR) clustering algorithm. WFR begins by computing resemblance scores between neighboring data instances based on resemblance function. A threshold is then applied to these scores to construct a resemblance graph, where edges indicate strong resemblance between points. The connected components of this graph naturally define clusters, without requiring prior knowledge of the number of clusters or assumptions about cluster shapes.

We evaluate the performance of WFR on a variety of toy benchmark datasets. Experimental results demonstrate that the algorithm effectively captures complex, nonlinear cluster structures. These results highlight the potential of WFR as a flexible, philosophically motivated approach to unsupervised learning, bridging insights from analytic philosophy and modern machine learning \cite{ghojogh2023philomatics}.

This paper is organized as follows. Section \ref{section_background} provides the background on Wittgenstein's family resemblance in analytic philosophy. The proposed WFR clustering is presented in Section \ref{section_WFR}. Section \ref{section_complexity} discusses the time and space complexities of the proposed algorithm. Simulations in Section \ref{section_simulations} justify the effectiveness of the proposed algorithm. Finally, Section \ref{section_conclusion} concludes the paper with a possible future direction for this algorithm. 

\section{Background on Wittgenstein's Family Resemblance in Analytic Philosophy}\label{section_background}

\subsection{Introduction to Ludwig Wittgenstein's Philosophy}

\subsubsection{Biography of Ludwig Wittgenstein}

Ludwig Wittgenstein (1889--1951) was an Austrian-British philosopher who made foundational contributions to logic, the philosophy of language, and the philosophy of mind. He studied engineering in Berlin and then moved to Cambridge to study under Bertrand Russell. Later, he became a professor of philosophy at the University of Cambridge. His work profoundly influenced analytic philosophy in the 20th century \cite{wittgenstein1990ludwig}.

\subsubsection{Early Wittgenstein (1911--1929)}

The early Wittgenstein, active primarily between 1911 and 1929, is exemplified by his work \textit{Tractatus Logico-Philosophicus}, published in 1921 \cite{wittgenstein1921tractatus}. In the \textit{Tractatus}, he develops a ``picture theory'' of language, according to which propositions represent the logical structure of reality and determine what can meaningfully be said. The work aims to draw sharp boundaries for language and dissolve philosophical problems by showing them as nonsensical. 
This book is highly structured, with a hierarchical, numbered system of propositions reminiscent of Spinoza's \textit{Ethics} \cite{spinoza1677ethics}, and it focuses on the logical form of language and its relation to the world \cite{stanford2002wittgenstein}.

\subsubsection{Later Wittgenstein (1930--1951)}

After a period away from philosophy, Wittgenstein returned and revised many of his earlier positions. His later philosophy, mainly from 1930 until his death in 1951, is most clearly presented in \textit{Philosophical Investigations}, published posthumously in 1953 \cite{wittgenstein1953philosophical}. In this book, he rejects his primary idea of a single universal logical structure and emphasizes that meaning arises from the use of language in context, introducing the notions of ``language-games'' and ``forms of life.'' Philosophy, in the later Wittgenstein, is therapeutic: it clarifies the actual use of language to resolve confusion rather than building a systematic theory \cite{stanford2002wittgenstein}.

\subsubsection{Key Differences of Early and Later Wittgenstein}

The main differences between early and later Wittgenstein can be summarized as follows:
\begin{itemize}
    \item \textbf{View of Language:} Early Wittgenstein sees language as picturing reality; later Wittgenstein sees language as diverse practices embedded in life.
    \item \textbf{Philosophical Aim:} Early work seeks a unified theory of meaning and limits of language; later work rejects grand theories and focuses on clarifying confusion in actual language use.
    \item \textbf{Method and Style:} The \textit{Tractatus} is formal and logical, whereas \textit{Investigations} is fragmentary and conversational.
    \item \textbf{Russell's View:} Bertrand Russell explicitly admired Wittgenstein’s early philosophy but strongly criticized his later philosophy \cite{russell1959my}.
\end{itemize}

\subsection{The Notion of Family Resemblance}

\begin{figure}[!t]
\centering
\includegraphics[width=2.5in]{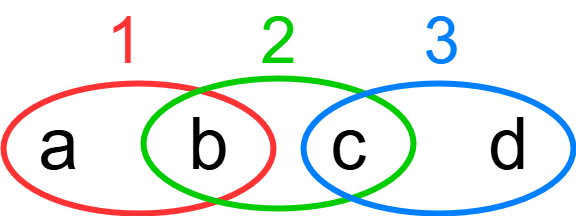}
\caption{Illustration of family resemblance: entity 1 possesses attributes (or so-called traits) $a$ and $b$, entity 2 possesses attributes $b$ and $c$, and entity 3 possesses attributes $c$ and $d$. All the entities 1, 2, and 3 belong to the same family, i.e., a chain of shared attributes, although entities 1 and 3 do not have any common attributes directly.}
\label{figure_family_resemblance}
\end{figure}

In \textit{Philosophical Investigations} \cite{wittgenstein1953philosophical}, \S\S 66--71\footnote{It is a standard scholarly shorthand, especially in philosophy and law. It means from section 66 through section 71 (inclusive). Note that, in philosophical manuscripts, \S\, and \S\S\, refer to Section and Sections, respectively.}, the later Wittgenstein introduced the notion of \textit{family resemblance}. To illustrate this idea, consider three entities, labeled 1, 2, and 3, as shown in Fig. \ref{figure_family_resemblance}. Suppose entity 1 possesses attributes (or so-called traits) $a$ and $b$, entity 2 possesses attributes $b$ and $c$, and entity 3 possesses attributes $c$ and $d$. Although entities 1 and 3 share no attributes directly, all three may nonetheless be regarded as belonging to the same family. This is because entity 1 shares an attribute with entity 2, and entity 2 shares an attribute with entity 3, forming a chain of overlapping similarities. Thus, resemblance is established not through a single common feature shared by all entities, but through a network or tree-like structure of partial overlaps.

\subsection{Use of Family Resemblance in Analytic Philosophy}

Wittgenstein’s notion of \emph{family resemblance} is widely used in analytic philosophy to analyze concepts that resist definition in terms of necessary and sufficient conditions. Instead of a single shared essence, such concepts are unified by overlapping similarities.
In the following, we review some of the use cases of family resemblance in analytic philosophy. 

\subsubsection{Ordinary Language Concepts}

Family resemblance is invoked when analyzing ordinary language concepts
whose extensions are heterogeneous.
Wittgenstein’s canonical example is the concept of a \emph{game} \citep[\S\S 66--71]{wittgenstein1953philosophical}. Board games, card games, athletic competitions, and solitary amusements share no single common feature. Instead, they exhibit overlapping similarities
such as rule-following, competition, skill, or entertainment.

\subsubsection{Philosophy of Language}

In analytic philosophy of language, family resemblance is used to support the thesis that meaning is grounded in use rather than in strict definitions \citep[\S 43]{wittgenstein1953philosophical}.
Words such as ``language'', ``sentence'', or ``meaning'' do not admit sharp boundaries but are understood through patterns of use across contexts; thus, they can be understood by family resemblance \citep[\S\S 65--67]{wittgenstein1953philosophical}.

\subsubsection{Philosophy of Science}

Family resemblance has been invoked in discussions of scientific classification, particularly in cases where categories resist essentialist definition \cite{dupre1993disorder,boyd1999homeostasis}. 
A prominent example arises in the classification of biological species. According to essentialist accounts of natural kinds, each genuine kind---such as a biological species or a chemical element---is characterized by an underlying essence, with membership determined by necessary and sufficient conditions \cite{boyd1999homeostasis}. However, empirical findings in biology reveal substantial variation among organisms within a species: no single genetic, morphological, or ecological trait is shared by all and only its members, and species boundaries are often indeterminate, as illustrated by phenomena such as ring species, hybridization, and asexual reproduction \cite{mayr1999systematics,dupre1993disorder}. In light of this, proponents of a family resemblance approach argue that species are unified not by a common essence but by overlapping clusters of traits, with different members sharing different subsets of these properties \cite{dupre1993disorder,boyd1999homeostasis}.

\subsubsection{Aesthetics and Philosophy of Art}

In analytic aesthetics, family resemblance is used to reject definitional theories of art. 
Paintings, musical works, performances, and conceptual art forms share overlapping similarities---such as expression, intention, and cultural role---without a single defining property \cite{weitz2017role,carroll2012philosophy}.

\subsubsection{Ethics and Moral Philosophy}

Family resemblance has been used in ethics and moral philosophy.
Moral concepts such as virtue, responsibility, or vice (opposite of virtue) are often treated as family resemblance concepts. For example, Courage, honesty, generosity, and kindness resemble one another without sharing a common essence \cite{murdoch2013sovereignty}.
This supports anti-reductive approaches to moral theory \cite{hacker2010virtue}.

\subsubsection{Philosophy of Mind}

In philosophy of mind and cognitive science, family resemblance is used to analyze mental categories such as emotions or intelligence.
Emotions like fear, anger, joy, and shame lack a single shared physiological or intentional structure, instead forming a cluster of related phenomena \cite{ryle2009concept,griffiths2008emotions}.

\section{Wittgenstein's Family Resemblance Clustering Algorithm}\label{section_WFR}

We propose the Wittgenstein's Family Resemblance (WFR) clustering algorithm, which is inspired by family resemblance in philosophy. 
The details of this algorithm are explained in the following. 

\subsection{Main Idea}

\begin{figure}[!t]
\centering
\includegraphics[width=3in]{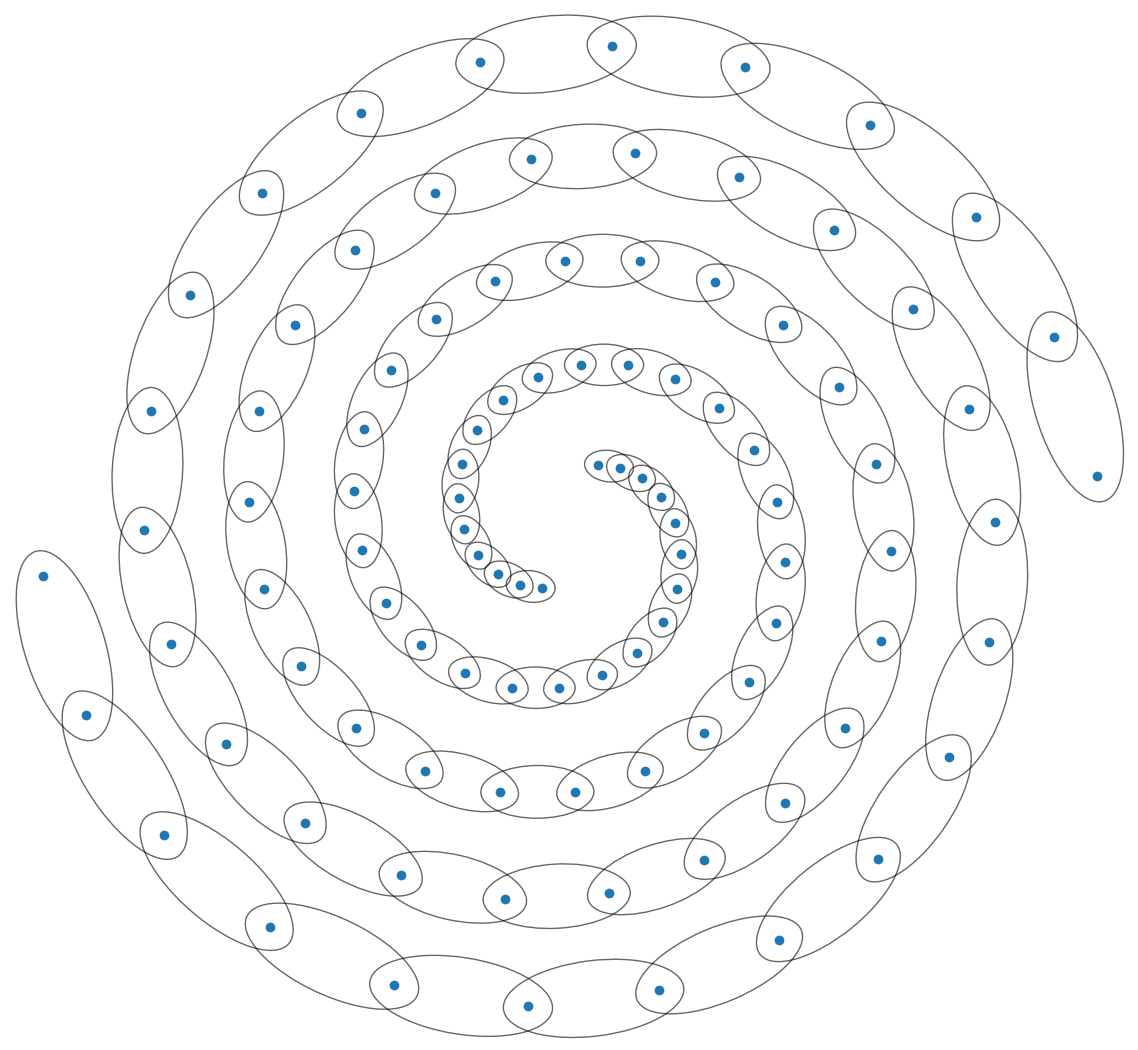}
\caption{Illustration of the main idea of WFR clustering algorithm, where the similar neighboring data instances form a family (or cluster) as a chain of similarities.}
\label{figure_main_idea}
\end{figure}

The central idea of WFR clustering is to compute pairwise resemblances between neighboring data instances. After applying a threshold to these resemblance values, instances with sufficiently high resemblance are linked, forming chains of similarity. These chains collectively define a family, which constitutes a cluster.
This behavior is illustrated in Fig. \ref{figure_main_idea} where two nonlinear clusters are found as two chains of similarities. 

\subsection{Resemblance Functions}

For forming the clusters, i.e., families of resemblance, a notion of resemblance should be defined to calculate the resemblance (or similarities) of the neighboring data instances. 
Various resemblance functions can be used as long as they are increasing functions with respect to similarities. 

The resemblance function is a map:
\begin{equation}
\begin{aligned}
&r: \mathcal{X} \times \mathcal{X} \rightarrow \mathbb{R}, \\
&r: \b{x}_1, \b{x}_2 \mapsto r(\b{x}_1, \b{x}_2),
\end{aligned}
\end{equation}
where $\mathcal{X}$ is the space of data and $\b{x}_1$ and $\b{x}_2$ are two data instances. 
The $r(\b{x}_1, \b{x}_2) \in \mathbb{R}$ denotes the resemblance score of data instances $\b{x}_1 \in \mathbb{R}^d$ and $\b{x}_2 \in \mathbb{R}^d$. 
Some example resemblance functions, for calculating the resemblance scores, are introduced in the following. 

\subsubsection{Log-based Resemblance Function}

The log-based resemblance function is:
\begin{align}
r(\b{x}_1, \b{x}_2) := \frac{1}{1 + \log\!\big(\|\b{x}_1 - \b{x}_2\|_2 + 1 + \epsilon\big)},
\end{align}
where $\|.\|_2$ denotes the $\ell_2$ norm and $\epsilon$ is a small positive number for numerical stability. This resemblance score is in range $[0, 1]$ where dissimilar and equal data instances have resemblances $0$ and $1$, respectively.

\begin{figure*}[!t]
\centering
\includegraphics[width=6.5in]{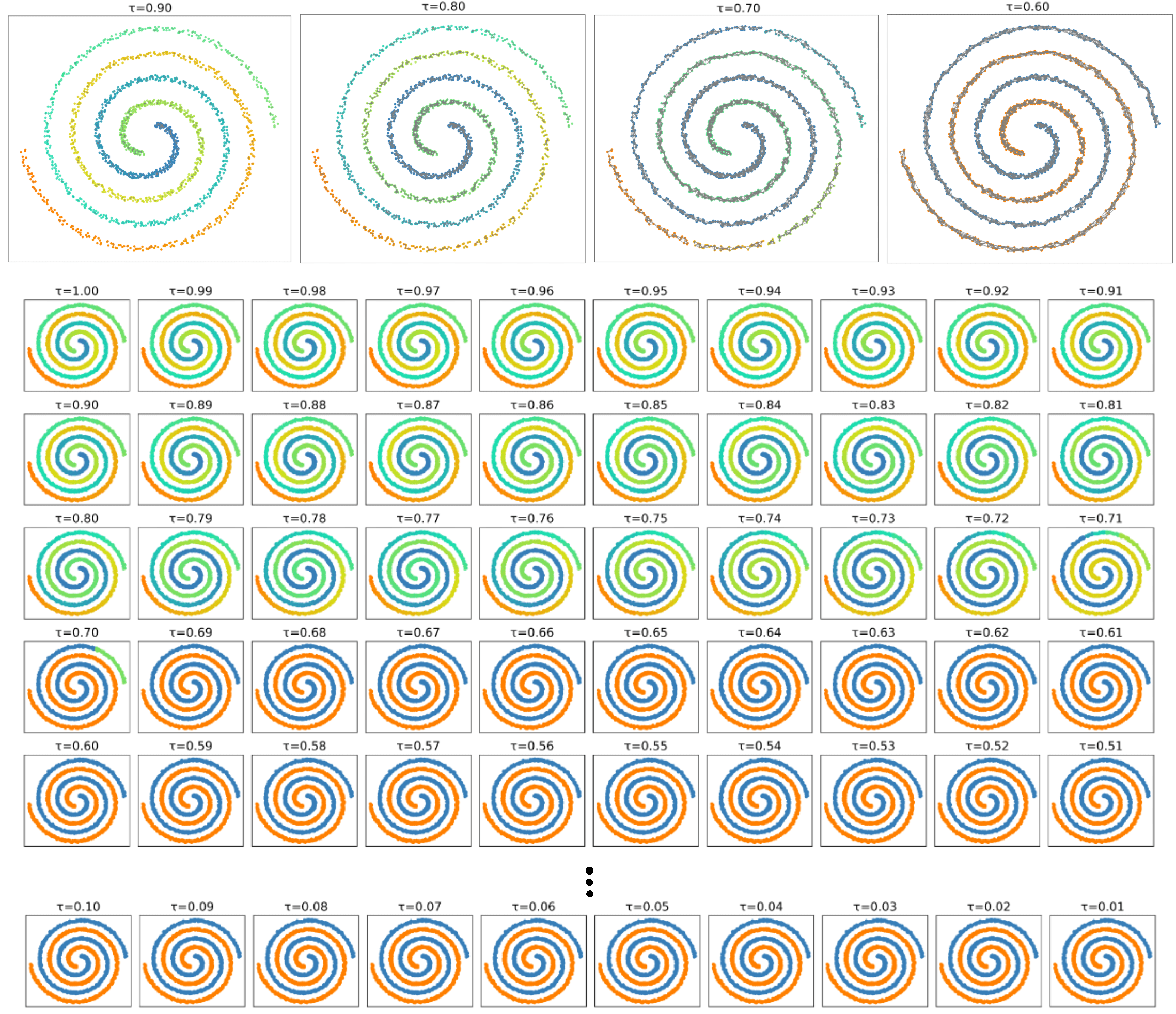}
\caption{(Top) Setting $\tau=0.90$ produces many clusters, while $\tau=0.80$ yields fewer. At $\tau=0.70$, the number of clusters decreases to six due to existing gaps between some data instances, and finally at $\tau=0.60$, the clustering correctly identifies two clusters. (Bottom) The grid search for the optimal threshold $\tau$ on the two-spirals dataset. Beginning the search at $\tau=1$ produces an excessive number of clusters. As the threshold is decreased, the number of clusters is gradually reduced. For thresholds $\tau \leq 0.69$, the combined score $s_1 + s_2$ improves, and the algorithm correctly identifies the two underlying clusters.}
\label{figure_toy_example4_log_based}
\end{figure*}

\subsubsection{Cosine Resemblance Function}

The cosine resemblance function is:
\begin{align}
r(\b{x}_1, \b{x}_2) := \cos(\b{x}_1, \b{x}_2) = \frac{\b{x}_1^\top \b{x}_2}{\|\b{x}_1\|_2\, \|\b{x}_2\|_2},
\end{align}
which is the inner product of normalized data instances. 
This resemblance score is in range $[-1, 1]$ where dissimilar and equal data instances have resemblances $-1$ and $1$, respectively. The cosine function calculates the resemblance in terms of angles between data instances. 

\subsubsection{Kernel Resemblance Function}

Kernel functions can be used for resemblance (or similarity) measurements. 
The kernel resemblance function is:
\begin{align}
r(\b{x}_1, \b{x}_2) := k(\b{x}_1, \b{x}_2) = \phi(\b{x}_1)^\top \phi(\b{x}_2),
\end{align}
where $k(.,.)$ is the kernel function and $\phi(.)$ is the pulling function from the input space to the Reproducing Kernel Hilbert Space (RKHS) \cite{ghojogh2023background}.

Any kernel function, such as Radial Basis Function (RBF) and sigmoid kernels, can be used for the kernel resemblance. If the kernel resemblance function is used in WFR clustering, the algorithm can be named \textit{kernel WFR} clustering algorithm. 

\subsection{Thresholding Resemblances and Search for Family Resemblances}


\subsubsection{Nearest Neighbors Graph}

We first find the $k$-Nearest Neighbors ($k$NN) of the training data instances. Thus, the neighbors of each data instance is found and a $k$NN graph is formed. 
Different algorithms can be used for finding the $k$NN graph \cite{bhatia2010survey}.
Some of the possible algorithms to use are brute-force (exhaustive) nearest neighbor, KD-Tree nearest neighbor \cite{friedman1977algorithm}, and Ball Tree nearest neighbor \cite{omohundro1989five,dolatshah2015ball}, which are implemented in Scikit-learn library \cite{pedregosa2011scikit}. 

\subsubsection{Resemblance Calculation and Thresholding}

Using the resemblance function, we calculate the resemblance of each data instance with its $k$-NN. 
This creates a resemblance graph, represented by a resemblance matrix $\b{R} \in \mathbb{R}^{n \times n}$ where $n$ denotes the number of training data instances. If the $i$-th instance has the $j$-th instance as its neighbor, the $(i,j)$-th element of the resemblance matrix is:
\begin{align}\label{equation_resemblance_matrix}
\b{R}_{ij} := 
\begin{cases}
r(\b{x}_i, \b{x}_j) & \mbox{if } \b{x}_j \in k\text{NN}(\b{x}_i) \\
0 & \mbox{Otherwise,}
\end{cases}
\end{align}
where $\b{R}_{ij}$ denotes the $(i,j)$-th element of the matrix $\b{R}$, and the resemblance scores can optionally be transformed to be non-negative. 

The resemblance matrix is normalized to be between zero and one:
\begin{align}
\widehat{\b{R}} := \frac{1}{r_{\text{max}} - r_{\text{min}}} (\b{R} - r_{\text{min}}),
\end{align}
where $\text{min}$ and $\text{max}$ are the minimum and maximum resemblances in the resemblance matrix. 

Then, a threshold $\tau \in [0, 1]$ is applied to the normalized resemblances to wipe out the weak resemblances below the threshold.
We define the (possibly asymmetric) adjacency matrix $\b{A} \in \{0,1\}^{n \times n}$ as:
\begin{equation}
\b{A}_{ij} :=
\begin{cases}
1, & \text{if } \widehat{\b{R}}_{ij} \geq \tau \\
0, & \text{otherwise}.
\end{cases}
\end{equation}

Since $\b{A}$ may be asymmetric, because of using $k$NN with $k < n$), we enforce symmetry by defining the final adjacency matrix:
\begin{equation}
\widetilde{\b{A}} = \b{A} \lor \b{A}^\top,
\end{equation}
where $\lor$ denotes the elementwise logical OR.
Equivalently, this can be written entrywise as:
\begin{equation}
\widetilde{\b{A}}_{ij} = \max\{\b{A}_{ij}, \b{A}_{ji}\}.
\end{equation}
Note that we use logical OR rather than mutual kNN to avoid disconnecting chains of resemblance.

To ensure the presence of self-loops, we set the diagonal entries of the adjacency matrix to one:
\begin{equation}
\widetilde{\b{A}}_{ii} = 1, \quad \forall i \in \{ 1,\dots,n \}.
\end{equation}

\subsubsection{Search in the Adjacency Graph}

The adjacency matrix $\widetilde{\b{A}}$ induces an adjacency graph in which adjacent data instances are connected. Graph search algorithms \cite{cormen2022introduction}, such as Depth-First Search (DFS) \cite{tarjan1972depth} or Breadth-First Search (BFS) \cite{moore1959shortest}, can be employed to identify the connected components of this graph. Each connected component corresponds to a family, or cluster, in which data instances are related through a chain of resemblance.
Binary connectivity captures the existence of family resemblance chains rather than their strength.

\subsection{Optional Outlier Marking}

Outlier detection can be optionally incorporated into WFR clustering by treating data instances belonging to small clusters as outliers with label $-1$. One criterion for identifying small clusters is to label clusters whose size is below a fixed ratio (e.g., $0.05$) of the maximum cluster size. Alternatively, a statistical approach can be adopted by fitting a normal distribution to the cluster sizes and designating clusters with sizes smaller than the mean minus a specified multiple of the standard deviation as small clusters. 

\subsection{Automatic Thresholding}

Clustering is inherently an ill-defined problem, as the perceived number of clusters can vary depending on the observer’s perspective. For instance, one person may identify two clusters in a dataset, while another may perceive three clusters by examining finer separations. Consequently, all clustering algorithms involve at least one hyperparameter that determines or influences the number of clusters. Some algorithms, such as K-means \cite{mcqueen1967some}, explicitly require the number of clusters as input, whereas others, such as DBSCAN \cite{ester1996density}, include hyperparameters that indirectly affect the number of clusters.

Similarly, WFR incorporates the threshold $\tau$ as its key hyperparameter, which controls the number of resulting clusters. Higher values of $\tau$ impose stricter thresholding, leading to a larger number of clusters. As illustrated in Fig. \ref{figure_toy_example4_log_based} (top), setting $\tau=0.90$ produces many clusters, while $\tau=0.80$ yields fewer. At $\tau=0.70$, the number of clusters decreases to six due to existing gaps between some data instances, and finally at $\tau=0.60$, the clustering correctly identifies two clusters.

However, thresholding in the WFR clustering algorithm can be performed automatically, albeit at the cost of increased computational time during the clustering phase\footnote{This procedure increases computational cost linearly with the number of candidate thresholds.}. This can be achieved by defining a grid of candidate thresholds within the range $[0, 1]$, using a fixed step size (e.g., $0.01$). For each candidate threshold, clusters are determined according to the procedure described previously. Subsequently, the following evaluation scores are computed.

The graph-based cluster separation score is defined as
\begin{align}\label{equation_s1}
s_1 = 1 - \frac{ \displaystyle \sum_{i=1}^{n} \sum_{j \in \mathcal{N}_k(i)} \frac{ \mathbb{I}[\ell_i \neq \ell_j] }{ d_{ij} + \varepsilon } }{ \displaystyle \sum_{i=1}^{n} \sum_{j \in \mathcal{N}_k(i)} \frac{1}{d_{ij} + \varepsilon} },
\end{align}
where $\ell_i \in \{1, \dots, c\}$ is the cluster label of $\b{x}_i$, $\mathcal{N}_k(i)$ is the set of $k$ nearest neighbors of $\b{x}_i$, $d_{ij} = \|\b{x}_i - \b{x}_j\|_2$ is the Euclidean distance between $\b{x}_i$ and $\b{x}_j$, $\varepsilon > 0$ is a small constant for numerical stability, and $\mathbb{I}[\ell_i \neq \ell_j]$ is the indicator function that equals 1 if $\b{x}_i$ and $\b{x}_j$ are in different clusters, and 0 otherwise.

The cluster size score is defined as:
\begin{equation}\label{equation_s2}
\begin{aligned}
s_2 = &\left( \frac{1}{c} \sum_{j=1}^{c} \min \Big( \frac{f_j}{f_{\min}}, 1 \Big) \right) \times \\
&~~~~~~~~~~~ \exp\!\Big(\! - \alpha \, \mathrm{Var}(f_1, \dots, f_c) \Big),
\end{aligned}
\end{equation}
where $c$ is the number of clusters, $n_j$ is the number of points in cluster $j$, $n$ is the total number of points, $f_j = n_j / n$ is the fraction of points in cluster $j$, $f_{\min}$ is the minimum acceptable cluster fraction (e.g., $0.05$), $\alpha \ge 1$ (e.g., $2.0$) is the imbalance penalty strength, and $\mathrm{Var}(f_1, \dots, f_c)$ is the variance of cluster fractions across all clusters.

The score $s_1$ increases when clusters are well-separated, whereas $s_2$ increases when cluster sizes are not excessively small. Both scores lie within the range $[0, 1]$. The optimal threshold $\tau$ for WFR clustering is the one that maximizes the sum of these scores:
\begin{align}
\tau := \arg\max_{\tau}\, (s_1 + s_2).
\end{align}
Note that it is also possible to use alternative clustering evaluation scores, such as Silhouette score \cite{rousseeuw1987silhouettes} or Davies--Bouldin index \cite{davies2009cluster}.

Figure \ref{figure_toy_example4_log_based} (bottom) illustrates the grid search for the optimal threshold $\tau$ on the two-spirals dataset. Beginning the search at $\tau=1$ produces an excessive number of clusters. As the threshold is decreased, the number of clusters is gradually reduced. For thresholds $\tau \leq 0.69$, the combined score $s_1 + s_2$ improves, and the algorithm correctly identifies the two underlying clusters.

\subsection{Test Phase (Out-of-sample Clustering)}

In the test phase for out-of-sample clustering, the $k$ Nearest Neighbors ($k$NN) of each test data instance are first identified among the training data instances. The resemblance scores between each test data instance and its $k$NN training instances are then computed. This process yields the test resemblance graph, represented by the test resemblance matrix $\b{R}' \in \mathbb{R}^{n_t \times n}$, where $n$ and $n_t$ denote the numbers of training and test samples, respectively. Let $\b{x}'_i$ denote the $i$-th test data instance and $\b{x}_j$ denote the $j$-th training data instance. The $(i,j)$-th entry of the test resemblance matrix is defined as:
\begin{align}\label{equation_test_resemblance_matrix}
\b{R}'_{ij} :=
\begin{cases}
r(\b{x}'_i, \b{x}_j), & \text{if } \b{x}_j \in k\text{NN}(\b{x}'_i), \\
0, & \text{otherwise}.
\end{cases}
\end{align}

The test resemblance matrix is normalized using the minimum and maximum values of the training resemblance matrix:
\begin{align}
\widehat{\b{R}}' := \frac{1}{r_{\text{max}} - r_{\text{min}}} (\b{R}' - r_{\text{min}}).
\end{align}
This ensures consistency between training and test resemblance scales.
Any entries of the normalized matrix that fall below zero or exceed one are clipped to zero and one, respectively.

For each test data instance, if the maximum normalized resemblance to its $k$NN exceeds the threshold $\tau$ (or the optimal threshold obtained via automatic thresholding), the cluster label of the training data instance with the highest resemblance is assigned to the test instance:
\begin{equation}\label{equation_test_label}
\ell'_{i} :=
\begin{cases}
\ell_j, & \text{if } \max_j \widehat{\b{R}}'_{ij} \geq \tau \\
-1, & \text{otherwise},
\end{cases}
\end{equation}
where $\ell'_i$ denotes the cluster label of the $i$-th test data instance and $\ell_j$ denotes the cluster label of the $j$-th training data instance. As shown in Eq. (\ref{equation_test_label}), if the maximum normalized resemblance falls below the threshold, the test instance is considered as an outlier and assigned the label $-1$.

\begin{figure}[!t]
\centering
\includegraphics[width=3.3in]{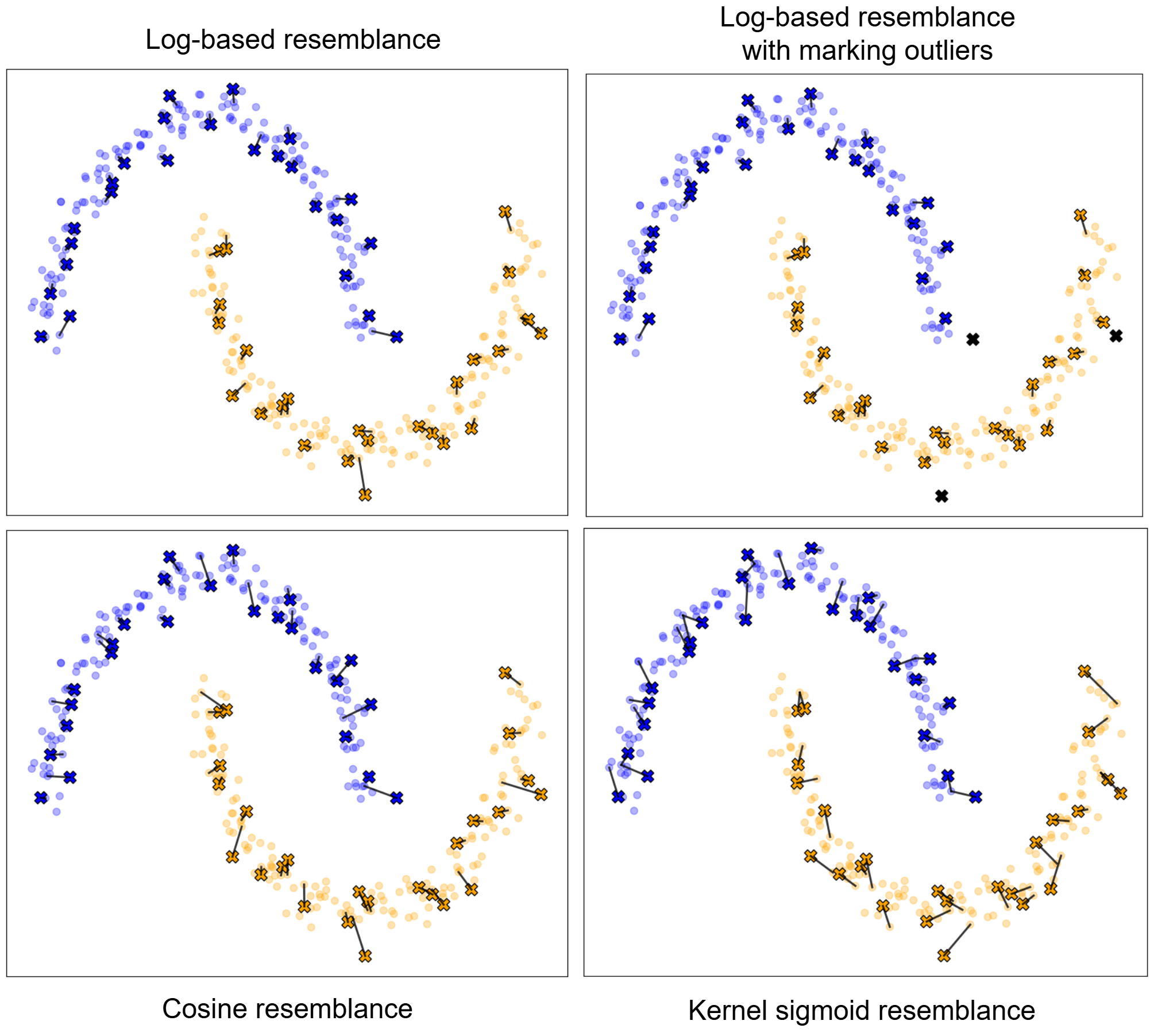}
\caption{Illustration of the test phase in WFR clustering with various resemblance functions. The lines show the largest resemblance between each test instance and its most similar training instance.}
\label{figure_test_phase}
\end{figure}

The test phase of WFR clustering, using log-based, cosine, and sigmoid kernel resemblance functions, is illustrated in Fig. \ref{figure_test_phase}. In this figure, test data instances are marked by crosses, with colors indicating their assigned cluster labels. The black lines connect each test instance to its neighboring training instance with the highest resemblance score.

\subsection{Algorithms for Training and Test Phases}

The algorithms for training and test phases of the proposed WFR clustering are shown in Algorithms \ref{algorithm_WFR_training} and \ref{algorithm_WFR_test}, respectively. 

In training, as in Algorithm \ref{algorithm_WFR_training}, the $k$NN graph of training data is calculated. Then, the resemblance matrix is computed and normalized. If automatic thresholding is used, the best threshold is detected by a grid search. If automatic thresholding is used, the hyperparameter $\tau$ is used as the threshold. Finally, the adjacency graph is calculated and a DFS or BFS search is performed in the adjacency graph. 

In the test phase, as in Algorithm \ref{algorithm_WFR_test}, the $k$NN graph of test data is computed among the training data. Then, the test resemblance matrix is calculated and normalized using the minimum and maximum of the training resemblance matrix. Finally, the label of each test data instance is determined if its maximum resemblance with a neighbor training data instance is above the threshold. Otherwise, it is an outlier. 

\SetKwProg{Fn}{Function}{:}{EndFunction}
\SetAlCapSkip{0.5em}
\IncMargin{0.8em}
\begin{algorithm2e}[!t]
\DontPrintSemicolon
    \textbf{Input}: training dataset $\{\b{x}_i\}_{i=1}^n$.\;
    \textbf{Output}: cluster labels $\{\ell_i\}_{i=1}^n$.\;
    \;
    Calculate the $k$NN graph of training data.\;
    $\b{R}_{ij} = 
    \begin{cases}
    r(\b{x}_i, \b{x}_j) & \mbox{if } \b{x}_j \in k\text{NN}(\b{x}_i) \\
    0 & \mbox{Otherwise}
    \end{cases}
    $\;
    $\widehat{\b{R}} = \frac{1}{r_{\text{max}} - r_{\text{min}}} (\b{R} - r_{\text{min}})
    $\;
    \If{automatic\_thresholding}{
        \For{$\tau \textbf{ from } \text{1 to 0 with fine steps}$}{
            $\{\ell_i\}_{i=1}^n \gets \text{Search}(\widehat{\b{R}}, \tau, \text{outlier\_marking})$
        }
        $s_1 \gets$ Eq. (\ref{equation_s1})\;
        $s_2 \gets$ Eq. (\ref{equation_s2})\;
        $\tau = \arg\max_{\tau}\, (s_1 + s_2)$\;
    }
    $\{\ell_i\}_{i=1}^n \gets \text{Search}(\widehat{\b{R}}, \tau, \text{outlier\_marking})$\;
    \KwRet $\{\ell_i\}_{i=1}^n$\;
    
    \;
    
    \Fn{$\text{Search}(\widehat{\b{R}}, \tau, \text{outlier\_marking})$}{
      $\b{A}_{ij} :=
        \begin{cases}
        1, & \text{if } \widehat{\b{R}}_{ij} \geq \tau \\
        0, & \text{otherwise}
        \end{cases}
        $\;
        $\widetilde{\b{A}} = \b{A} \lor \b{A}^\top$\;
        $\widetilde{\b{A}}_{ii} = 1, \quad \forall i \in \{ 1,\dots,n \}$\;
        $\{\ell_i\}_{i=1}^n \gets$ Do DFS or BFS search in the graph $\widetilde{\b{A}}$.\;
        \If{outlier\_marking}{
            // mark tiny clusters as outliers:\;
            $\ell_j \gets -1$, for tiny clusters
        }
      \KwRet $\{\ell_i\}_{i=1}^n$\;
    }
    
\caption{The algorithm of training phase in WFR clustering.}\label{algorithm_WFR_training}
\end{algorithm2e}
\DecMargin{0.8em}

\SetAlCapSkip{0.5em}
\IncMargin{0.8em}
\begin{algorithm2e}[!t]
\DontPrintSemicolon
    \textbf{Input}: test dataset $\{\b{x}'_i\}_{i=1}^{n_t}$.\;
    \textbf{Output}: cluster labels $\{\ell'_i\}_{i=1}^{n_t}$.\;
    \;
    Calculate the $k$NN graph of test data among training data.\;
    $\b{R}'_{ij} :=
    \begin{cases}
    r(\b{x}'_i, \b{x}_j), & \text{if } \b{x}_j \in k\text{NN}(\b{x}'_i) \\
    0, & \text{otherwise}
    \end{cases}
    $\;
    $\widehat{\b{R}}' := \frac{1}{r_{\text{max}} - r_{\text{min}}} (\b{R}' - r_{\text{min}})
    $\;
    $\ell'_{i} :=
    \begin{cases}
    \ell_j, & \text{if } \max_j \widehat{\b{R}}'_{ij} \geq \tau \\
    -1, & \text{otherwise}
    \end{cases}
    $\;
    \KwRet $\{\ell'_i\}_{i=1}^{n_t}$\;
    
\caption{The algorithm of test phase in WFR clustering.}\label{algorithm_WFR_test}
\end{algorithm2e}
\DecMargin{0.8em}

\section{Time and Space Complexities}\label{section_complexity}

In this section, we analyze the time and space complexities of the proposed WFR clustering algorithm for both the training and test phases. Let $n$ denote the number of training samples, $n_t$ the number of test samples, $d$ the data dimensionality, and $k$ the number of nearest neighbors.

\subsection{The Complexity of Training Phase}

\paragraph{$k$NN Graph Construction:}
The first step of the training phase is constructing the $k$-Nearest Neighbors ($k$NN) graph for the training data.
Using a brute-force search, this step requires
$\mathcal{O}(n^2 d)$ time.
When spatial indexing structures such as KD-trees or Ball trees are applicable, the expected time complexity reduces to
$\mathcal{O}(n \log n \cdot d)$ for low- to moderate-dimensional data.
The space complexity for storing the kNN graph is $\mathcal{O}(nk)$.

\paragraph{Resemblance Computation and Normalization:}
Resemblance scores are computed only between each data instance and its $k$ nearest neighbors.
Therefore, resemblance computation requires
$\mathcal{O}(nk \cdot d)$ time, assuming that evaluating the resemblance function is linear in terms of the dimension $d$.
The resemblance matrix is sparse and requires $\mathcal{O}(nk)$ memory.
Normalization of the resemblance values requires a single pass over the nonzero entries and thus has time complexity $\mathcal{O}(nk)$.

\paragraph{Thresholding and Graph Construction:}
Applying the threshold to the resemblance matrix and enforcing symmetry both require $\mathcal{O}(nk)$ time and space, as only $k$ neighbors per node are considered.
The resulting adjacency graph remains sparse.

\paragraph{Graph Search for Connected Components:}
Identifying connected components using DFS or BFS has time complexity
$\mathcal{O}(n + |E|)$, where $|E| = \mathcal{O}(nk)$ is the number of edges in the adjacency graph.
Thus, this step requires $\mathcal{O}(nk)$ time and $\mathcal{O}(n)$ additional space for bookkeeping.

\paragraph{Automatic Thresholding (Optional):}
If automatic thresholding is employed using a grid search over $T$ candidate threshold values, the graph construction and search steps are repeated $T$ times.
In this case, the total training time complexity becomes:
\begin{align}
\mathcal{O}\big(n^2 d + (T \cdot nk) \big),
\end{align}
for brute-force kNN, or:
\begin{align}
\mathcal{O}\big((n \log n \cdot d) + (T \cdot nk) \big),
\end{align}
when using efficient nearest neighbor search.
The space complexity remains $\mathcal{O}(nk)$.
Note that $T$ is a fixed number so it can be ignored in the complexity analysis. 

\subsection{The Complexity of Test Phase}

In the test phase, each test data instance finds its $k$ nearest neighbors among the training data.
Using brute-force search, this requires $\mathcal{O}(n_t n d)$ time, which can be reduced to
$\mathcal{O}(n_t \log n \cdot d)$ using tree-based nearest neighbor methods.
Computing resemblance scores between test instances and their neighbors requires $\mathcal{O}(n_t k d)$ time and $\mathcal{O}(n_t k)$ space.

Assigning cluster labels based on the maximum resemblance score for each test instance is a linear operation with time complexity $\mathcal{O}(n_t k)$ and negligible additional memory overhead.
Therefore, the overall test-time complexity is dominated by nearest neighbor search.

\begin{table*}[t]
\centering
\caption{Time and space complexity comparison of WFR clustering with baseline clustering algorithms.
Here, $n$ is the number of samples, $d$ is the dimensionality, $c$ is the number of clusters, $k$ is the number of nearest neighbors, and $I$ denotes the number of iterations.\\}
\label{table_complexity_comparison}
\begin{tabular}{lcc}
\toprule
Algorithm & Time Complexity & Space Complexity \\
\midrule
K-means \cite{mcqueen1967some}
& $\mathcal{O}(n c d I)$ 
& $\mathcal{O}(nd)$ \\

Gaussian Mixture Models (EM) 
& $\mathcal{O}(n c d I)$ 
& $\mathcal{O}(nd)$ \\

Affinity Propagation \cite{frey2007clustering} 
& $\mathcal{O}(n^2 I)$ 
& $\mathcal{O}(n^2)$ \\

Mean Shift \cite{cheng1995mean} 
& $\mathcal{O}(n^2 I)$ 
& $\mathcal{O}(n)$ \\

Spectral Clustering \cite{ng2001spectral}
& $\mathcal{O}(n^3)$ 
& $\mathcal{O}(n^2)$ \\

Agglomerative Clustering (Hierarchical) 
& $\mathcal{O}(n^2 \log n)$ 
& $\mathcal{O}(n^2)$ \\

Ward Clustering \cite{ward1963hierarchical}
& $\mathcal{O}(n^2)$ 
& $\mathcal{O}(n^2)$ \\

DBSCAN \cite{ester1996density}
& $\mathcal{O}(n \log n)$ 
& $\mathcal{O}(n)$ \\

OPTICS \cite{ankerst1999optics}
& $\mathcal{O}(n \log n)$ 
& $\mathcal{O}(n)$ \\

HDBSCAN \cite{campello2015hierarchical}
& $\mathcal{O}(n \log n)$ 
& $\mathcal{O}(n)$ \\

BIRCH \cite{zhang1996birch}
& $\mathcal{O}(n)$ 
& $\mathcal{O}(n)$ \\

\textbf{WFR (ours)} 
& $\mathcal{O}(n \log n \cdot d + nk)$ 
& $\mathcal{O}(nk)$ \\
\bottomrule
\end{tabular}
\end{table*}

\subsection{Comparison with Baseline Clustering Algorithms}

Table~\ref{table_complexity_comparison} compares the computational complexity of WFR clustering with several well-known clustering algorithms.
As shown in this table, many classical clustering algorithms incur quadratic or cubic time and space complexity, which limits their scalability to large datasets.
Spectral clustering \cite{ng2001spectral,shi2000normalized,ghojogh2023laplacian}, affinity propagation \cite{frey2007clustering}, and hierarchical methods \cite{murtagh2012algorithms,johnson1967hierarchical} require $\mathcal{O}(n^2)$ memory, while spectral clustering additionally requires $\mathcal{O}(n^3)$ time due to eigen-decomposition.

Density-based methods such as DBSCAN \cite{ester1996density}, OPTICS \cite{ankerst1999optics}, and HDBSCAN \cite{campello2015hierarchical} achieve near-linear time complexity under suitable indexing assumptions, but rely on density estimation and distance thresholds that may be sensitive to data distribution and scaling.
Centroid-based methods such as K-means \cite{mcqueen1967some} and Gaussian mixture models \cite{mclachlan2000finite,dempster1977maximum,ghojogh2019fitting} scale well but require the number of clusters to be specified in advance and struggle with nonconvex cluster shapes.

The proposed WFR clustering operates on a sparse kNN resemblance graph and avoids dense similarity matrices.
Its time and space complexities scale linearly with respect to the number of edges in the graph, $\mathcal{O}(nk)$, making it comparable to density-based methods in scalability while retaining flexibility in similarity definition and supporting kernel-based and non-metric resemblances.
Furthermore, unlike many baseline methods, WFR naturally supports out-of-sample clustering.


\section{Simulations}\label{section_simulations}

\subsection{Comparison of Resemblance Graphs for Different Resemblance Functions}

\begin{figure}[!t]
\centering
\includegraphics[width=3.2in]{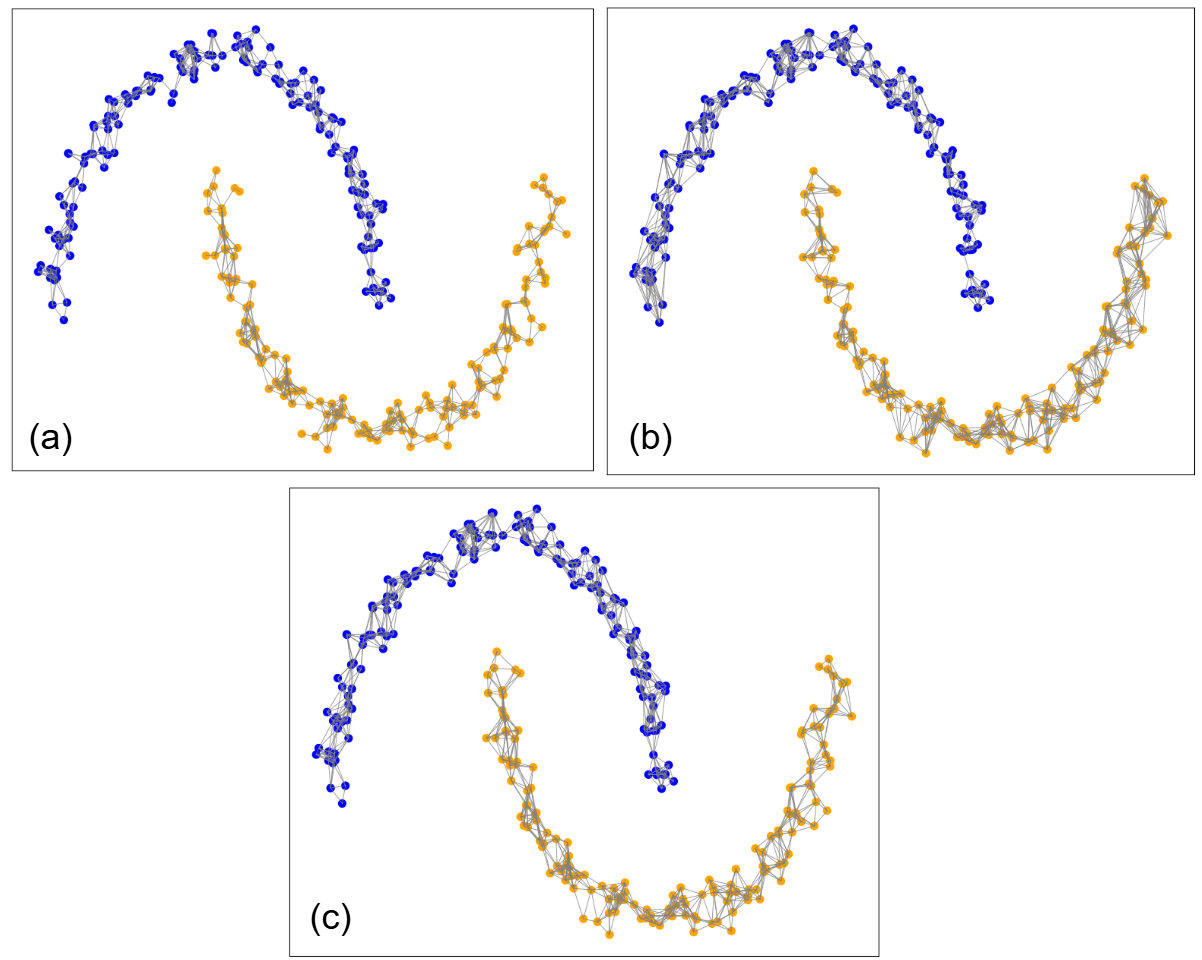}
\caption{Illustration of the adjacency graph for (a) log-based, (b) cosine, and (c) kernel RBF resemblance functions in the two-moons dataset. The gray lines show the thresholded resemblances between the training data instances.}
\label{figure_resemblance_graph}
\end{figure}

Figure \ref{figure_resemblance_graph} illustrates the adjacency graphs produced by the WFR algorithm using log-based, cosine, and RBF kernel resemblance functions on the two-moons dataset. In this figure, the gray lines represent the thresholded resemblances between training data instances. As the figure shows, different resemblance functions induce slightly different adjacency graphs, which in turn lead to minor variations in the resulting clustering.

\subsection{Comparison with Other Clustering Algorithms}

\begin{figure*}[!t]
\centering
\includegraphics[angle=90, width=5in]{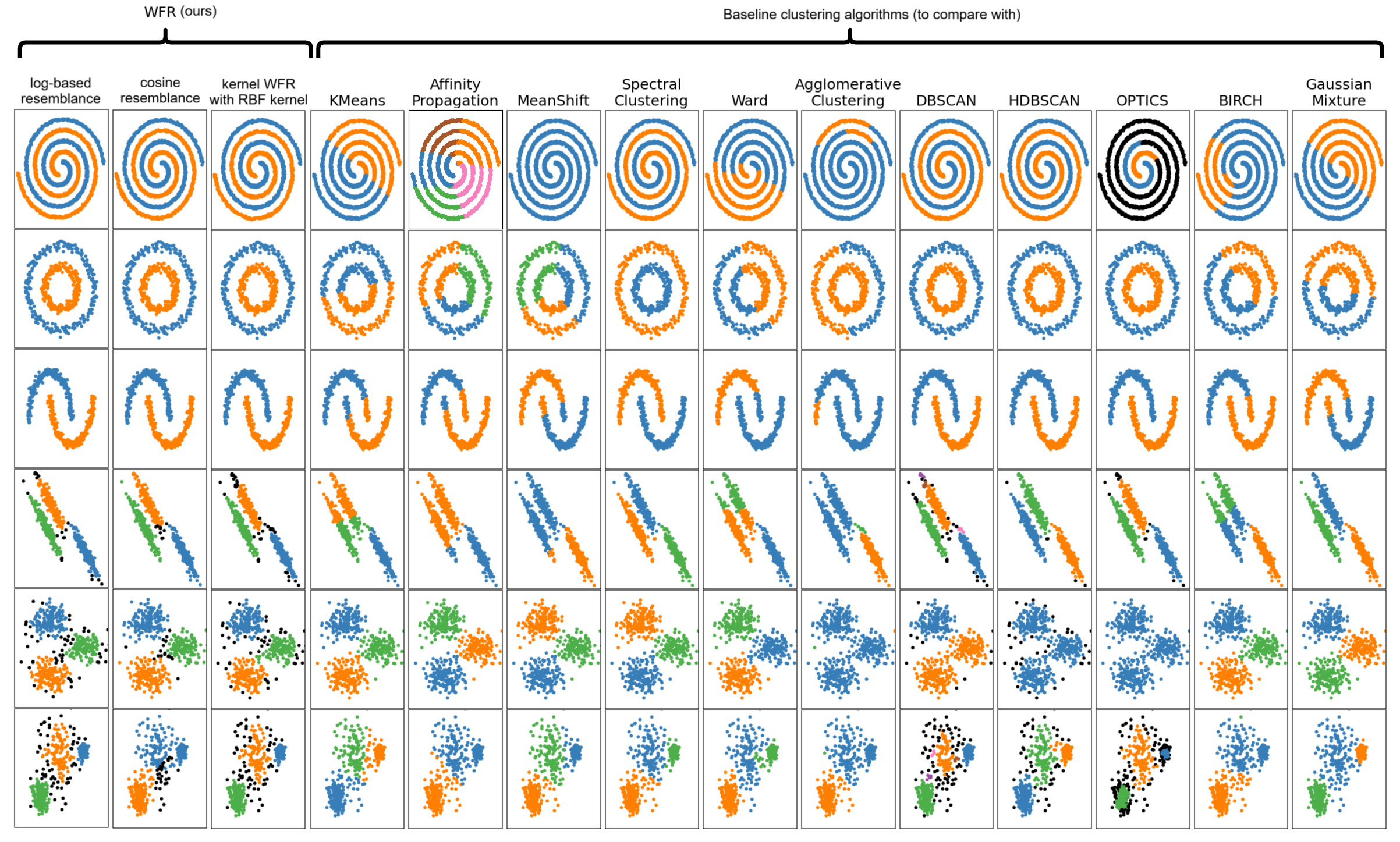}
\caption{Comparison of clustering algorithms on several benchmark datasets with diverse cluster shapes, obtained using different clustering algorithms. 
The benchmark datasets are two-spirals, two-circles, two-moons, and three additional benchmarks, each consisting of three Gaussian distributions with varying shapes.}
\label{figure_benchmarks}
\end{figure*}

In this section, we compare our proposed WFR clustering algorithm with well-known, effective clustering algorithms, including K-means \cite{mcqueen1967some}, affinity propagation \cite{frey2007clustering}, mean shift \cite{cheng1995mean,comaniciu2002mean}, spectral clustering \cite{ng2001spectral,shi2000normalized,ghojogh2023laplacian}, Ward \cite{ward1963hierarchical}, agglomerative clustering (hierarchical clustering) \cite{murtagh2012algorithms,johnson1967hierarchical}, DBSCAN \cite{ester1996density}, HDBSCAN \cite{campello2015hierarchical}, OPTICS \cite{ankerst1999optics}, BIRCH \cite{zhang1996birch}, and gaussian mixture models using Expectation Maximization (EM) \cite{mclachlan2000finite,dempster1977maximum,ghojogh2019fitting}. 
For the baseline clustering methods, we use their optimized libraries available in Scikit-learn \cite{pedregosa2011scikit}. 
A Scikit-learn–compatible implementation of the proposed WFR algorithm is available in the following GitHub repository: \url{https://github.com/bghojogh/WFR_Clustering}.

Figure \ref{figure_benchmarks} illustrates the clustering results on several benchmark datasets with diverse cluster shapes, obtained using different clustering algorithms. 
The benchmark datasets include two-spirals, two-circles, two-moons, and three additional benchmarks, each consisting of three Gaussian distributions with varying shapes.
As shown in the figure, the proposed WFR clustering method, employing various resemblance functions---namely log-based, cosine, and RBF kernel functions---performs effectively in identifying the correct cluster structures while accurately detecting outliers. In contrast, some baseline clustering algorithms fail to recover the true clusters, particularly in highly nonlinear scenarios. 
Unlike DBSCAN, WFR does not rely on density estimation or metric balls and supports arbitrary resemblance functions, including kernel-based similarities.
This shows the effectiveness of the proposed WFR and kernel WFR clustering algorithms in detecting true clusters in both linear and nonlinear scenarios.

\section{Conclusion and Future Directions}\label{section_conclusion}

This paper proposed a novel clustering algorithm, termed Wittgenstein’s Family Resemblance (WFR), along with its kernel-based variant. The algorithm is inspired by the concept of family resemblance introduced by Ludwig Wittgenstein in analytical philosophy and represents an instance of philomatics, wherein philosophical ideas are leveraged to develop mathematical methods.

In WFR, a nearest-neighbor graph is constructed, and resemblance (similarity) scores are computed between neighboring data instances. After applying a thresholding step, data instances that are connected through chains of resemblance are assigned to the same cluster. In its current formulation, the WFR algorithm employs a binary resemblance graph, where edges indicate only the presence or absence of resemblance.

Future work may extend WFR to weighted resemblance graphs, enabling the incorporation of resemblance strength while preserving the notion of family resemblance chains.
This extension would yield a weighted adjacency graph with thresholded resemblance values, allowing for more nuanced clustering behavior.


\section*{Acknowledgement}

The authors were unaware of the related work in \cite{xiao2010data} during the development of the WFR algorithm and the preparation of the manuscript. While related in spirit, their method differs from WFR clustering in several aspects. Interested readers may consult \cite{xiao2010data} for an alternative family-resemblance-based clustering approach. Connections between family resemblance and machine learning have also been discussed in informal venues \cite{philosophy2017stackexchange}. Family resemblance has further been explored in the context of fuzzy sets \cite{veri2023transforming}.


\bibliography{References}
\bibliographystyle{icml2016}

\end{document}